\documentclass{article}

\usepackage{microtype}
\usepackage{graphicx}
\usepackage{booktabs} % for professional tables

\usepackage{hyperref}

\usepackage[accepted]{icml2024}

\usepackage{amsmath}
\usepackage{amssymb}
\usepackage{mathtools}
\usepackage{amsthm}
\usepackage{subcaption}
\usepackage{enumitem}

\usepackage[capitalize,noabbrev]{cleveref}

\theoremstyle{plain}
\newtheorem{theorem}{Theorem}[section]

\theoremstyle{definition}
\newtheorem{definition}[theorem]{Definition}

\theoremstyle{remark}

\usepackage[textsize=tiny]{todonotes}

\icmltitlerunning{Denoising-based Contractive Imitation Learning}

\begin{document}

\twocolumn[
\title{Denoising-based Contractive Imitation Learning\\}



\icmlsetsymbol{equal}{*}

%\begin{icmlauthorlist}
%\icmlauthor{Anonymous authors}{yyy}
% \icmlauthor{Firstname2 Lastname2}{equal,yyy,comp}
% \icmlauthor{Firstname3 Lastname3}{comp}
%\end{icmlauthorlist}
\author{
    Macheng Shen$^{1}$, Jishen Peng$^{1,2}$, Zefang Huang$^{1,3}$ \\
    {\normalsize $^{1}$Shanghai Qi Zhi Institute, $^{2}$Shanghai Jiaotong University} \\
    {\normalsize $^{3}$Nanjing University}
}

%\icmlaffiliation{yyy}{Department of XXX, University of YYY, Location, Country}


\date{}
\maketitle
\vskip 0.3in
]




\begin{abstract}
A fundamental challenge in imitation learning is the \emph{covariate shift} problem. Existing methods to mitigate covariate shift often require additional expert interactions, access to environment dynamics, or complex adversarial training, which may not be practical in real-world applications.
In this paper, we propose a simple yet effective method \textbf{DeCIL}, \textbf{De}noising-based \textbf{C}ontractive \textbf{I}mitation \textbf{L}earning, to mitigate covariate shift by incorporating a denoising mechanism that enhances the contraction properties of the state transition mapping. Our approach involves training two neural networks: a dynamics model \( f \) that predicts the next state from the current state, and a joint state-action denoising policy network \( d \) that refines this state prediction via denoising and outputs the corresponding action. We provide theoretical analysis showing that the denoising network acts as a local contraction mapping, reducing the error propagation of the state transition and improving stability. Our method is simple to implement and integrates seamlessly with existing imitation learning frameworks without requiring additional expert data or complex modifications to the training procedure. Empirical results demonstrate that our approach effectively improves success rate of various imitation learning tasks under noise perturbation. Code can be viewed in \url{https://github.com/MachengShen/Stable-BC}.

\end{abstract}

\section{Introduction}

Imitation learning enables agents to acquire complex behaviors by learning from expert demonstrations \citep{pomerleau1989alvinn, argall2009survey}. It has been successfully applied in robotics \citep{billard2008robot}, autonomous driving \citep{bojarski2016end}, and game playing \citep{silver2016mastering}. However, a fundamental challenge in imitation learning is the \emph{covariate shift} problem \citep{ross2010efficient, ross2011reduction}, where discrepancies between the training and execution state distributions lead to compounding errors. The learned policy may encounter states during execution that were not represented in the training data, resulting in poor generalization and degraded performance.

Existing methods to mitigate covariate shift often require additional expert interaction, access to environment dynamics, or complex training procedures, which may not be practical in real-world applications.

In this paper, we propose a simple yet effective approach to mitigate covariate shift by enhancing the contraction properties of state transitions through a denoising mechanism. Our method involves training two neural networks:

\begin{enumerate}
    \item \textbf{Dynamics Model \( f \)}: Predicts the next state \( \hat{x}_{t+1} \) given the current state \( x_t \).
    \item \textbf{Denoising Policy Network \( d \)}: Takes the current state \( x_t \) and the predicted next state \( \hat{x}_{t+1} \) to output a refined next state \( \tilde{x}_{t+1} \) and the corresponding action \( \hat{a}_t \).
\end{enumerate}

Our key insight is that by incorporating a denoising step, we can reduce the Lipschitz constant of the state transition mapping, effectively making it a local contraction mapping. This reduces the impact of prediction errors, preventing them from compounding over time.

The main advantages of our approach are its simplicity and compatibility with existing imitation learning frameworks. It requires only access to expert demonstrations and does not necessitate additional expert interaction or complex training procedures. Moreover, our method can be easily integrated into standard training pipelines and can complement other techniques to further improve performance.

% Our contributions are:

% \begin{itemize}
%     \item We introduce a simple and effective method to address covariate shift in imitation learning by incorporating a denoising mechanism that enhances the contraction properties of state transitions.
%     \item We provide a theoretical analysis demonstrating that the denoising network increases the contraction of the state mapping, improving stability and preventing error accumulation.
%     \item We show that our approach is easily integrable with existing imitation learning methods and can serve as a building block for more complex algorithms.
%     \item We empirically validate our method on benchmark tasks, demonstrating that it outperforms baseline imitation learning algorithms in terms of stability and policy performance.
% \end{itemize}

\subsection{Paper Organization}

The rest of the paper is organized as follows: In Section~\ref{sec:related_works}, we discuss related works and position our approach in the context of existing methods. Section~\ref{sec:method} presents our method in detail. In Section~\ref{sec:theoretical_analysis}, we provide the theoretical analysis demonstrating the contraction properties of our approach. Section~\ref{sec:experiments} presents empirical results validating our method on benchmark tasks. Finally, we conclude in Section~\ref{sec:conclusion}.

\section{Related Works}
\label{sec:related_works}

Imitation learning aims to learn policies that mimic expert behavior using demonstration data \citep{argall2009survey}. Behavioral cloning (BC) \citep{pomerleau1989alvinn} treats imitation learning as supervised learning, training a policy to map states to actions directly from expert demonstrations. However, BC suffers from the covariate shift problem because the learned policy may encounter states during execution that are not represented in the training data, leading to compounding errors.

Several methods have been proposed to mitigate covariate shift:

\paragraph{Interactive Expert Queries.} Methods like DAgger \citep{ross2011reduction} and its variants \citep{laskey2017dart, sun2017deeply} involve querying the expert for corrective actions during the agent's own state distribution. While effective, these approaches require ongoing access to the expert, which may not be feasible in many practical scenarios.

\paragraph{Adversarial Imitation Learning.} Approaches such as GAIL \citep{ho2016generative} and AIRL \citep{fu2018learning} formulate imitation learning within a generative adversarial framework, where a discriminator distinguishes between expert and agent behaviors. These methods aim to align the agent’s trajectory distribution with that of the expert \citep{blonde2022lipschitzness_gail, ghasemipour2020divergence}. However, they often rely on querying the environment during training, which can be infeasible in many imitation learning scenarios where access to the environment is limited or costly. Additionally, adversarial training introduces challenges such as instability and sensitivity to hyperparameters \citep{wiatrak2019stabilizing}.

\paragraph{Data Augmentation.} Techniques that augment the training data have been explored to enhance robustness. For instance, \citet{ke2021imitation, jiang2024recovering} propose generating synthetic data using learned dynamics models, while \citet{florence2019self, zhou2023nerf, spencer2021feedback, hoque2024intervengen} leverage domain-specific invariances to create augmented samples. However, these methods often rely on additional assumptions, such as knowledge of system invariances or access to accurate dynamics models, which may not be available.

\paragraph{Stability and Contractive Policies.} Incorporating stability properties into policy learning has gained attention as a way to enhance robustness. \citet{blocher2017learning} and \citet{ravichandar2017learning} focus on learning stable dynamical systems that guarantee convergence to desired states. \citet{beik2024neuralCDS, abyaneh2024contractive} introduce contractive dynamical systems to ensure exponential convergence and improve out-of-sample recovery. While these methods provide theoretical guarantees, they often involve complex constrained optimization, specific architectures or assume accessibility to additional low-level controller, which can limit their scalability and practicality.

\paragraph{Simple and Integrable Approaches.} Recent works have highlighted the need for methods that are simple to implement and can be easily integrated with existing frameworks. \citep{ke2021grasping} propose a simple noise-perturbing mechanism to alleviate covariate-shift. \citep{mehta2024stable} propose Stable-BC, which reguluarizes the eigenvalues of the Jacobian of the closed-loop dynamics to achieve stability. However, their approach requires an accurate dynamics model to avoid trading-off performance for stability. Besides, computing Jacobian is intractable for high-dimensional state space, which restricts the applicability of their approach. 

\paragraph{Our Contribution.} In contrast to these approaches, our method provides a simple yet effective solution to covariate shift by incorporating a denoising mechanism. By training a denoising policy network with a denoising objective, we encourage contraction in the state transition mapping without requiring complex constraints, additional expert interaction, or access to environment dynamics. Our approach is easy to implement, requires minimal additional assumptions beyond access to expert data, and can be seamlessly integrated into existing imitation learning pipelines.

\section{Method}
\label{sec:method}
\subsection{Problem Formulation}
We consider the standard imitation learning setting where an agent aims to learn a policy $\pi: \mathcal{X} \rightarrow \mathcal{A}$ that maps states $x_t \in \mathcal{X}$ to actions $a_t \in \mathcal{A}$, based on expert demonstrations. The expert provides a dataset of trajectories consisting of tuples $(x_t, a_t, x_{t+1})$, where $x_{t+1}$ is the state resulting from taking action $a_t$ in state $x_t$.

\subsection{Continuous Perspective and Intuition}
\begin{figure}[t]
    \centering
    \includegraphics[width=0.85\columnwidth]{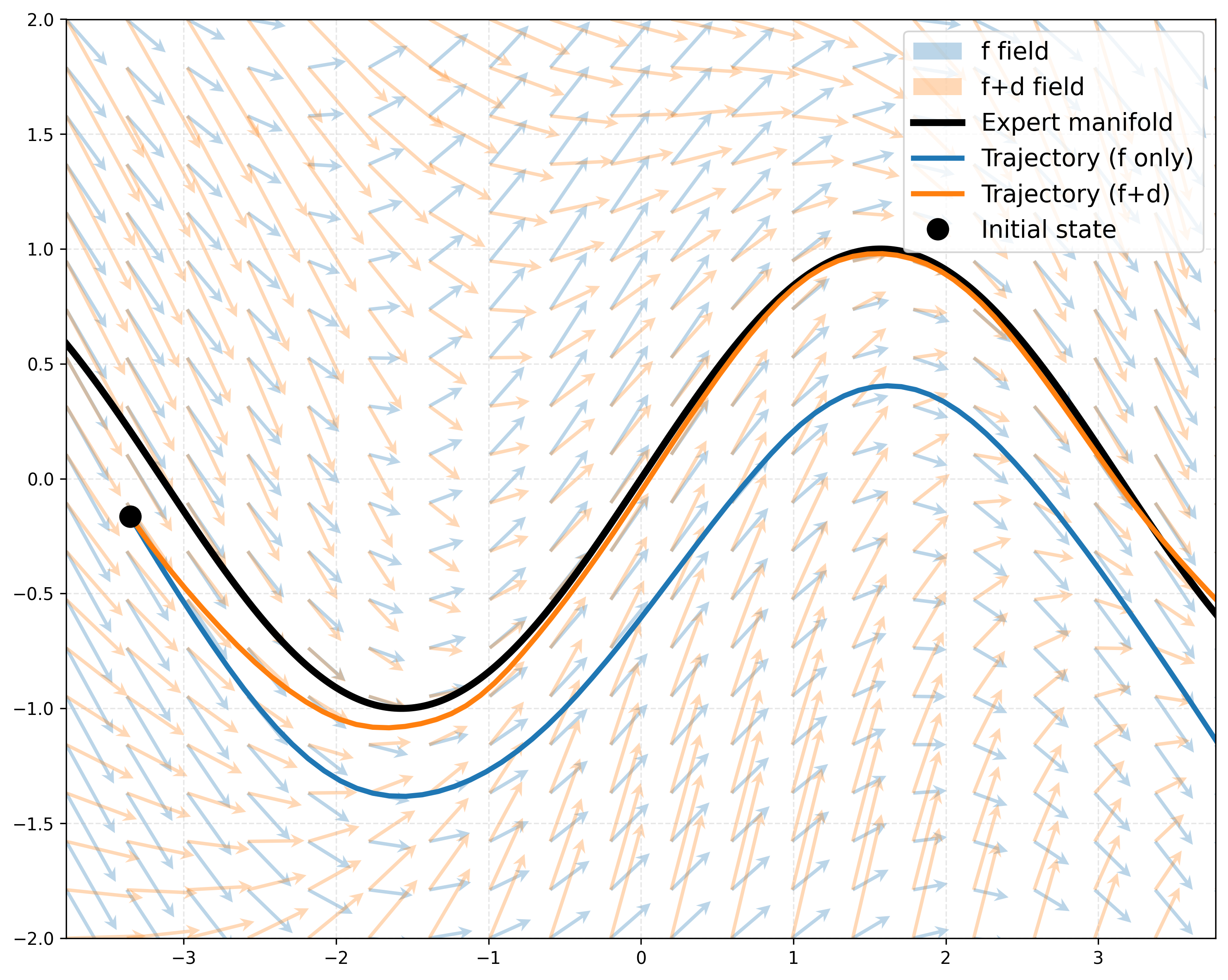}
    \caption{Comparison of trajectory prediction methods. The black curve shows the expert manifold (ground truth), and the black dot indicates a noisy initial state. The blue trajectory and vector field show the prediction using only the learned drift network $f$, while the orange trajectory and vector field show the prediction using the combined drift and denoising networks ($f + d$). The denoising network helps pull the trajectory back to the expert manifold, effectively preventing covariate-shift.}
    \label{fig:traj_comparison}
\end{figure}

To gain intuition behind our proposed approach, we will first consider a continuous-time dynamical system modeling the evolution of a state $x(t) \in \mathcal{X}$:
\begin{equation}
\label{eq:continuous_dynamics}
\frac{dx}{dt} = f(x) + d(x),
\end{equation}
where $f(x)$ describes the nominal state transition dynamics, and $d(x)$ represents an additional term that biases the trajectory towards regions of high data density. One way to choose $d(x)$ is to relate it to the \emph{score function}, defined as $d(x) = \nabla \log p(x)$, where $p(x)$ is the data distribution implied by the expert demonstrations.

Intuitively, if $p(x)$ is high (i.e., $x$ lies on or near the expert trajectory distribution), then $\nabla \log p(x)$ guides the trajectory to remain close to these high-probability regions. Conversely, if the trajectory begins to deviate towards low-density areas, the score function $d(x)$ points it back towards states consistent with the training data, mitigating deviations that would otherwise accumulate over time, as illustrated in Fig. \ref{fig:traj_comparison}. In other words, $d(x)$ acts as a stabilizing force that counteracts the inherent drift caused by model imperfections or noise.

In this continuous view, ensuring that $d(x)$ effectively contracts the state space towards the data manifold can prevent compounding errors and improve the overall stability of the learned policy. However, directly implementing such a continuous mechanism in imitation learning can be challenging, especially when only discrete samples of expert demonstrations are available and when the dynamics must be approximated from data rather than having explicit access to $f(x)$ or $p(x)$.

Our proposed method can be seen as a discrete approximation of this continuous perspective: we learn a discrete-time policy that incorporates a denoising mechanism analogous to the score function $d(x)$. By refining state predictions and actions at each timestep, our approach approximates the continuous guidance of trajectories back to the data manifold.

\subsection{Overview of the Approach}

Our method involves training two neural networks:
\begin{enumerate}
    \item \textbf{Dynamics Model $f$}: Predicts the next state $\tilde{x}_{t+1}$ from the current state $x_t$.
    \item \textbf{Denoising Policy Network $d$}: Takes the current state $x_t$ and a potentially noisy next state $\tilde{x}_{t+1}$ to output a refined next state $\hat{x}_{t+1}$ and the corresponding action $\hat{a}_t$.
\end{enumerate}

During training, $f$ approximates the state transition dynamics, while $d$ is trained via a denoising objective that encourages it to correct prediction errors and prevent the trajectory from drifting away from the training distribution. The interplay between $f$ and $d$ serves as a discrete-time approximation of the continuous-time dynamics: $f$ provides a nominal prediction, and $d$ acts like a score-based correction term, pushing the system back towards states consistent with the expert data.

In the following sections, we detail how $f$ and $d$ are trained, provide theoretical analysis showing that $d$ induces contraction in the state transition, and present empirical results demonstrating that our approach outperforms baseline methods in terms of stability and robustness to noise.

\subsection{Training the Dynamics Model $f$}
The dynamics model $f$ is trained to minimize the mean squared error (MSE) between the predicted next state and the true next state:
\begin{equation}
\label{eq:loss_f}
\mathcal{L}_f = \mathbb{E}_{(x_t, x_{t+1})} \left[ \left\| f(x_t) - x_{t+1} \right\|^2 \right].
\end{equation}

\subsection{Training the Denoising Policy Network $d$}
The denoising network $d$ is trained to map a noisy next state back to the true next state and predict the corresponding action. The input to $d$ is the concatenation of the current state $x_t$ and a noisy version of the next state $y$, where:
\[
y = x_{t+1} + \eta,
\]
and $\eta$ is noise sampled from a distribution $\mathcal{N}(0, \sigma^2 I)$.

The outputs of $d$ are:
\[
[ \hat{x}_{t+1}, \hat{a}_t ] = d( x_t, y ).
\]

\subsubsection{Denoising Objective}
The denoising loss encourages $d$ to reconstruct the true next state $x_{t+1}$ from the noisy input and predict the action jointly:
\begin{align}
\label{eq:loss_denoise}
\mathcal{L}_{d} &= 
\mathcal{L}_{\text{denoise}} + \lambda \mathcal{L}_{\text{action}} \nonumber \\
&=
\mathbb{E}_{(x_t, a_t, x_{t+1}), \eta} \left[ \left\| \hat{x}_{t+1} - x_{t+1} \right\|^2 + \lambda \left\| \hat{a}_t - a_t \right\|^2 \right],
\end{align}

where $\lambda$ balances the importance of the two terms.

\subsection{Inference Procedure}
During inference, the agent performs the following steps:
\begin{enumerate}
    \item \textbf{Predict the Next State Using $f$}:
    \begin{equation}
    \label{eq:infer_f}
    \tilde{x}_{t+1} = f(x_t).
    \end{equation}
    \item \textbf{Refine the Prediction and Generate the Action with $d$}:
    \begin{equation}
    \label{eq:infer_g}
    [\hat{x}_{t+1}, \hat{a}_t] = d\left( x_t, \tilde{x}_{t+1} \right).
    \end{equation}
    \item \textbf{Execute the Predicted Action}:
    \begin{equation}
    \label{eq:infer_env}
    x_{t+1}' = D\left( x_t, \hat{a}_t \right),
    \end{equation}
    where $D$ represents the environment's dynamics function.
\end{enumerate}

By refining the predicted next state, $d$ corrects potential errors introduced by $f$, resulting in a more stable state transition and an accurate action prediction.

\subsection{Theoretical Analysis}
\label{sec:theoretical_analysis}

We provide theoretical justification to show that incorporating the denoising network $d$ enhances the contraction of the mapping from $x_t$ to $x_{t+1}'$, compared to using the dynamics model $f$ alone. We first show the enhanced contraction from \( x_t \) to \( \hat{x}_{t+1} \), and then show that the environment transition preserves this property. 

\subsubsection{Jacobian-Based Error Propagation Analysis}
\label{sec:theoretical_analysis}

We analyze how the composite mapping from \( x_t \) to \( \hat{x}_{t+1} \) affects error propagation by examining the Jacobian of the composite function. Let:
\[
h(x_t) = g(x_t, f(x_t)) = \hat{x}_{t+1}.
\]

Hereafter, we use \( g \) to denote the sub-mapping \( (x_t, \tilde{x}_{t+1}) \rightarrow \hat{x}_{t+1} \) through the denoising network \( d \). Consider a reference trajectory \( x_t^* \) (e.g., the expert trajectory) and define the error:
\[
e_t = x_t - x_t^*.
\]

For small \( e_t \), the error propagation follows:
\begin{equation}
\label{eq:error_propagation}
e_{t+1} \approx J_h(x_t^*) e_t,
\end{equation}
where \( J_h(x_t^*) = \frac{\partial h}{\partial x}\big|_{x_t^*} \) is the Jacobian of \( h \) at \( x_t^* \).

\subsubsection{Decomposition of the Composite Jacobian}

Since \( h(x) = g(x, f(x)) \), the chain rule gives:
\[
J_h(x_t^*) = \frac{\partial g}{\partial x}(x_t^*, f(x_t^*)) + \frac{\partial g}{\partial y}(x_t^*, f(x_t^*)) J_f(x_t^*),
\]
where \( J_f(x_t^*) = \frac{\partial f}{\partial x}\big|_{x_t^*} \) and \( y = f(x_t) \) is the predicted next state.

Define:
\[
J_{g,x} = \frac{\partial g}{\partial x}(x_t^*, f(x_t^*)) \quad \text{and} \quad J_{g,y} = \frac{\partial g}{\partial y}(x_t^*, f(x_t^*)).
\]

Thus:
\begin{equation}
\label{eq:h_jacobian}
J_h(x_t^*) = J_{g,x} + J_{g,y} J_f(x_t^*).
\end{equation}

Here, \( J_f(x_t^*) \) captures how perturbations in \( x_t \) affect the next state prediction \( f(x_t) \). Without correction, \( f \) might cause trajectories to drift away from the data manifold, amplifying errors.

\subsubsection{Role of the Denoising Network and the Residual Interpretation}

The denoising network \( g(x_t, y) \) refines the predicted next state \( y = f(x_t) \) and outputs an action. Intuitively, \( g \) is trained to reduce noise in \( y \), preventing drift from the expert trajectory distribution.

Formally, we can view \( g \) as performing a residual correction:
\[
\hat{x}_{t+1} = y - \epsilon(x_t, y),
\]
where \( \epsilon(x_t, y) \) represents the learned noise estimation. For small noise, \( \epsilon(x_t, y) \) is small, and its partial derivatives with respect to \( x_t \) are also small. This suggests that \( J_{g,x} \) is small since \( g \)'s corrections depend less on the input state \( x_t \) and more on the predicted next state \( y \). In other words, \( g \) does not rely heavily on \( x_t \) to refine the state, thus limiting the sensitivity captured by \( J_{g,x} \).

On the other hand, ensuring a small \(J_{g,y}\) is nontrivial. However, by training $g$ with a denoising objective, we effectively penalize its sensitivity to noise in the input $y$. In Appendix~\ref{appendix:proof}, we show that when $g$ is trained to correct noisy samples of \(x_{t+1}\) the gradient of $g$ with respect to $y$ \(\bigl( \text{i.e., }J_{g,y}\bigr)\) is pushed to be small in norm. Concretely, the denoising loss can be viewed as minimizing \(\|J_{g,y}\|\), by penalizing how much small changes in $y$ affect the output. Hence, training under noise naturally drives $g$ to exhibit a lower Lipschitz constant with respect to $y$, thereby curbing error amplification through the predicted next state channel.

\subsubsection{Mitigating Drift and Enhancing Contraction}

Combining the above:
\[
J_h(x_t^*) = J_{g,x} + J_{g,y} J_f(x_t^*).
\]

\begin{itemize}
    \item If \( J_f(x_t^*) \) tends to increase errors, the presence of \( g \) can counteract this effect by introducing corrections that limit error growth.
    \item With \( J_{g,x} \) small due to the residual interpretation and \( J_{g,y} \) shown to be small in the appendix, the composite mapping \( h \) is less prone to error amplification.
    \item While we may not guarantee strict contraction (i.e., \(\|J_h(x_t^*)\|<1\) for all \(x_t^*\)) under all conditions, the presence of \( g \) reduces the effective Jacobian norm of \( h \), thereby mitigating drift and pushing the system closer to a regime where errors do not explode over time.
\end{itemize}

In essence, the denoising network \( g \) provides a correction mechanism that reduces the system's sensitivity to both \( x_t \) and \( y \) perturbations. Even if strict contraction is not guaranteed, this mechanism helps maintain trajectories near the data manifold, mitigating the compounding errors associated with covariate shift.

\subsubsection{Justification for the Error Bound in the Environment Dynamics}

An important aspect of our approach is that during inference, the agent executes the predicted action $\hat{a}_t$ in the environment, resulting in the next state $x_{t+1}' = D(x_t, \hat{a}_t)$. To ensure that the contraction property holds when interacting with the actual environment, we need to show that $x_{t+1}'$ is close to the refined predicted next state $\hat{x}_{t+1}$, and that the error incurred is bounded by the losses minimized during training.

Our training data consists of tuples $(x_t, x_{t+1}, a_t)$ collected from expert demonstrations, where $x_{t+1} = D(x_t, a_t)$. The denoising policy network $g$ is trained to minimize both the denoising loss $\mathcal{L}_{\text{denoise}}$ and the action prediction loss $\mathcal{L}_{\text{action}}$, ensuring that:

\begin{equation}
\label{eq:denoise_loss_small}
\| \hat{x}_{t+1} - x_{t+1} \| \leq \epsilon_x,
\end{equation}
\begin{equation}
\label{eq:action_loss_small}
\| \hat{a}_t - a_t \| \leq \epsilon_a,
\end{equation}
where $\epsilon_x$ and $\epsilon_a$ are small constants representing the minimized losses.

\paragraph{Error Propagation Through Environment Dynamics}

While global Lipschitz continuity of the environment dynamics $D$ may not always hold, it is often reasonable to assume that, within the region of state-action space explored by the policy, small variations in actions yield proportionally small changes in the resulting next states. Formally, if $D$ is locally Lipschitz continuous with respect to the action, with Lipschitz constant $L_D^a$, then:

\begin{equation}
\label{eq:env_lipschitz_action}
\| D(x_t, \hat{a}_t) - D(x_t, a_t) \| \leq L_D^a \| \hat{a}_t - a_t \| \leq L_D^a \epsilon_a.
\end{equation}

Since $x_{t+1} = D(x_t, a_t)$ and $x_{t+1}' = D(x_t, \hat{a}_t)$, it follows that:

\begin{equation}
\label{eq:state_difference}
\| x_{t+1}' - x_{t+1} \| \leq L_D^a \epsilon_a.
\end{equation}

\paragraph{Connecting $\hat{x}_{t+1}$ and $x_{t+1}'$}

Combining inequalities \eqref{eq:denoise_loss_small} and \eqref{eq:state_difference}, we can bound the difference between the refined predicted next state $\hat{x}_{t+1}$ and the actual next state $x_{t+1}'$ obtained by executing $\hat{a}_t$ in the environment:

\begin{align}
\| x_{t+1}' - \hat{x}_{t+1} \| &\leq \| x_{t+1}' - x_{t+1} \| + \| x_{t+1} - \hat{x}_{t+1} \| \nonumber \\
&\leq L_D^a \epsilon_a + \epsilon_x. \label{eq:total_error_bound}
\end{align}

This shows that the error between $x_{t+1}'$ and $\hat{x}_{t+1}$ is bounded by terms involving the minimized training losses $\epsilon_a$ and $\epsilon_x$.

\paragraph{Implications for Contraction Mapping}

Given that $\hat{x}_{t+1}$ is close to $x_{t+1}'$, and under the assumption that the composite mapping $h(x_t) = \hat{x}_{t+1}$ is contracting (as previously established), the overall mapping from $x_t$ to $x_{t+1}'$ via the environment dynamics and the predicted action remains close to a contraction mapping, with errors bounded by the training losses.

This justifies that executing the predicted action $\hat{a}_t$ in the environment does not significantly disrupt the contraction property established by the denoising network $d$. The small errors introduced are controlled by the minimized losses and the continuity of the environment dynamics, ensuring stability and preventing error accumulation over time.

%\bibliographystyle{plainnat}
%\bibliography{icml2024/example_paper}

\section{Experiments}
\label{sec:experiments}

In this section, we first provide empirical validation of the theoretical analysis presented in Section~\ref{sec:theoretical_analysis}. Specifically, we investigate how the \emph{sensitivity reduction ratio} varies with increasing noise factors, thereby demonstrating the resilience of our denoising-based approach against noise perturbations.

\begin{definition}[Sensitivity Ratio]
For a given behavior cloning (BC) model \( f \) trained to predict the next state \( x_{t+1} \) from the current state \( x_t \), and a composite model \( f\circ d \) where \( d \) is the denoising network (with a bit abuse of notation, here we use \( f\circ d \) to denote the process of first applying \( f \), then \( d \)), the \emph{sensitivity} \( S_f(x_t) \) and \( S_{f\circ d}(x_t) \) at state \( x_t \) are defined as follows:
\[
S_f(x_t) = \mathbb{E}_{\eta \sim \mathcal{N}(0, \sigma_s^2 I)} \left[ \left\| f(x_t + \eta) - x_{t+1} \right\| \right],
\]
\[
S_{f\circ d}(x_t) = \mathbb{E}_{\eta \sim \mathcal{N}(0, \sigma_s^2 I)} \left[ \left\| (f\circ d)(x_t + \eta) - x_{t+1} \right\| \right].
\]

Noting that $\sigma_s$ is a fixed small noise factor which is different from the noise factor $\sigma$ for training the denoising network.

The \emph{sensitivity reduction ratio} \( \rho(x_t) \) is then defined as:
\[
\rho(x_t) = \frac{S_{f\circ d}(x_t)}{S_f(x_t)}.
\]
\end{definition}

The sensitivity ratios \( \rho(x_t) \) are averaged over all states in the training dataset to obtain a mean sensitivity reduction ratio for each \( \sigma \).

\begin{figure}[t]
    \centering
    \includegraphics[width=0.95\columnwidth]{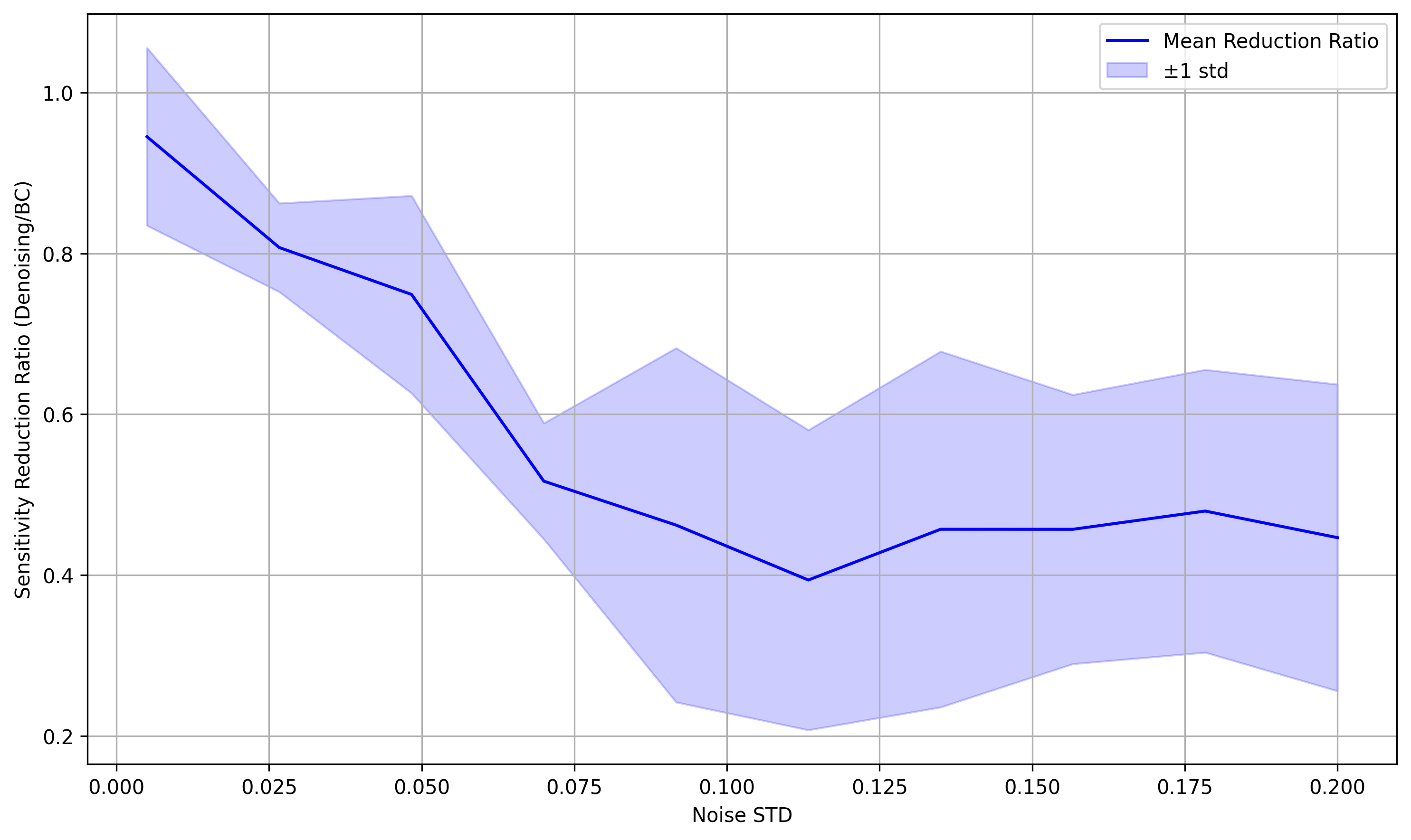}
    \vspace{-0.5em}
    \caption{\textbf{Sensitivity Reduction Ratio vs.\ Noise Factor.} 
    The plot illustrates how the sensitivity reduction ratio \( \rho \) changes with increasing Gaussian noise standard deviation \( \sigma \). A ratio \( \rho < 1 \) indicates that the denoising mechanism effectively reduces sensitivity compared to behavior cloning (BC) alone. The ratio reaches a minimum at around \( \sigma = 0.1 \), demonstrating optimal noise resilience. Beyond this point, the ratio increases, suggesting that excessive noise forces the denoising network to rely more heavily on the current state \( x_t \) to infer the next state, thereby diminishing the contraction effect. This behavior aligns with our residual interpretation, highlighting the efficacy of the denoising mechanism under moderate noise levels while indicating limitations when noise becomes too large.
    }
    \label{fig:sensitivity_noise}
\end{figure}

Fig.~\ref{fig:sensitivity_noise} presents the relationship between the noise factor \( \sigma \) and the sensitivity reduction ratio \( \rho \), on a simple sinusoidal curve dataset similar to the one shown in Fig.~\ref{fig:traj_comparison} but with discrete implementation. The mean reduction ratio is averaged over 3 random seeds for each noise factor. The results demonstrate that:

\begin{itemize}
    \item The sensitivity reduction ratio \( \rho \) is consistently below 1. This indicates that the denoising network \( d \) effectively mitigates the sensitivity introduced by the behavior cloning model \( f \), thereby enhancing the overall resilience of the state transition mapping against noise perturbations.
    \item At around \( \sigma = 0.1 \), the sensitivity reduction ratio \( \rho \) reaches its minimum value, showcasing the optimal performance of the denoising mechanism in reducing sensitivity.
    \item For higher noise levels (\( \sigma > 0.1 \)), the ratio \( \rho \) begins to increase again. This trend suggests that excessive noise overwhelms the denoising network's capacity to effectively contract the mapping, forcing it to infer the clean next state primarily from the current state \( x_t \) instead of the predicted next state. As a result, the contraction effect is weakened, and the sensitivity reduction is compromised. This result further complements our theoretical analysis provided under small noise assumption.
\end{itemize}

\subsection{Experimental Setup}
\label{sec:experimental_setup}

Next, we evaluate our method on two benchmark environments: the \textit{Intersection} environment from \citep{mehta2024stable} and the \textit{MetaWorld} environments from \citep{yu2020meta, ke2021imitation}. These environments test the robustness of imitation learning algorithms under noisy conditions. We compare our approach to the following baselines:

\begin{itemize}
    \item \textbf{Behavior Cloning (BC)}: A standard supervised learning approach that directly maps states to actions using expert demonstrations.
    \item \textbf{Diffusion policy} \citep{chi2023diffusion_policy}: An action distribution learning approach using diffusion model.
    \item \textbf{DART}: A data augmentation method that generates synthetic data to reduce covariate shift \citep{sun2017deeply}.
    \item \textbf{Stable-BC} \citep{mehta2024stable}: A method that regularizes the eigenvalues of the Jacobian of the closed-loop dynamics to achieve stability. % This baseline is evaluated on both intersection driving and quadrotor control environments.
    \item \textbf{Noisy BC}: A variant of behavior cloning where noise is added to the input states during training \citep{ke2021grasping}.
    \item \textbf{MOReL}: A method that trains an ensemble of dynamics functions and uses the variance between the model output as a proxy estimation of uncertainty to stay within high-confidence region. \citep{kidambi2020morel}.
\end{itemize}

\begin{figure*}[!h]
\centering
\begin{subfigure}[thbp]{0.48\textwidth}
\centering
\includegraphics[width=\linewidth]{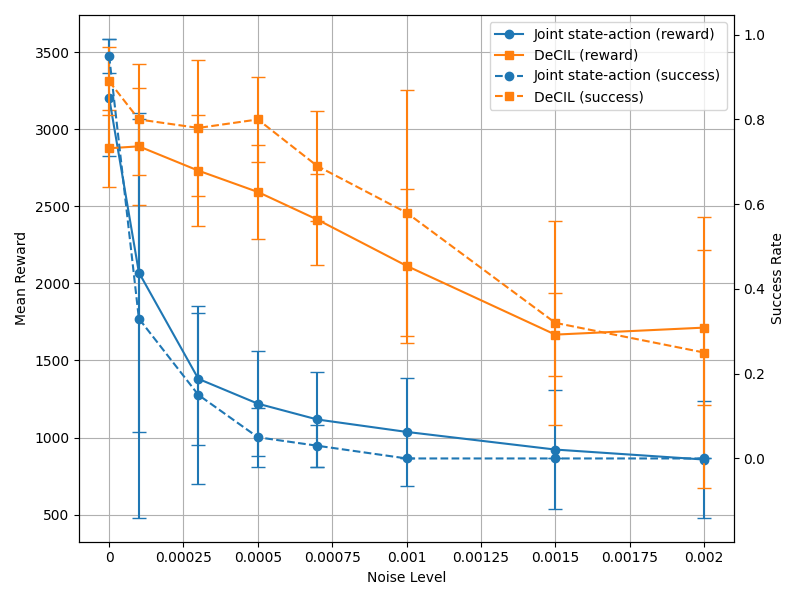}
\caption{Metaworld: Button Press}
\label{fig:button_press}
\end{subfigure}
\hfill
\begin{subfigure}[thbp]{0.48\textwidth}
\centering
\includegraphics[width=\linewidth]{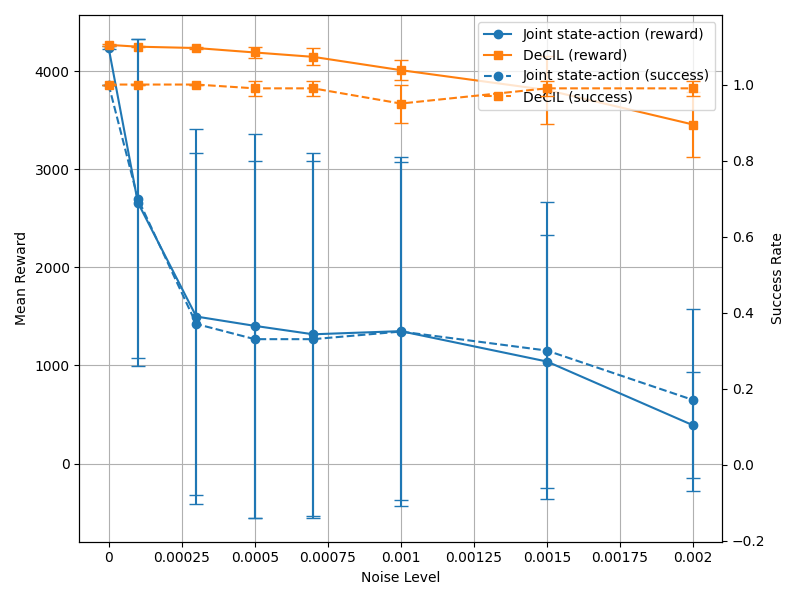}
\caption{Metaworld: Drawer Close}
\label{fig:drawer_close}
\end{subfigure}
\caption{Ablation study comparing DeCIL with a joint state-action prediction baseline. As noise increases, DeCIL retains high performance, while the baseline's performance degrades rapidly.}
\label{fig:noise_ablation}
\end{figure*}

\subsection{Environment Setup}

\subsubsection{Intersection Environment}
The Intersection environment simulates a robot navigating a dynamic intersection (as described in \cite{mehta2024stable}). 
We measure performance by a reward function defined as the \emph{negation} of the cost from \cite{mehta2024stable}.%:
% \[
%     \text{Reward}(x, o, c) = -\Bigl(
%       \|x(t + \Delta t) - c\| - \|x(t) - c\|
%       + 0.75 \cdot \|x(t) - o(t)\|
%       - 0.75 \cdot \|x(t + \Delta t) - o(t)\|
%     \Bigr),
% \]
% where $c$ is the constant goal position, $x(t)$ is the autonomous agent’s state, and $o(t)$ is the other vehicle’s state. 
% Larger reward thus indicates the agent is both moving closer to the goal and avoiding collisions.

We train each model using 2, 5, 10, 20, or 50 trajectories, and evaluate under two scenarios:
\begin{enumerate}[label=\textbullet]
    \item Case 1: The test environment matches the training environment exactly.
    \item Case 2: The simulated human (other vehicle) is self-centered and only reasons about its own state, creating a state distribution shift.
\end{enumerate}

\subsubsection{MetaWorld Environments}
The MetaWorld environments \cite{yu2020meta} consist of diverse robotic manipulation tasks, each requiring complex object manipulation given expert trajectories. 
We evaluate on a subset of MetaWorld tasks used in \cite{ke_ccil}, using 10 demonstrations for training and varying the levels of state noise. 
These tasks are higher-dimensional than the Intersection environment, posing additional challenges for methods that rely on Jacobian-based regularization (e.g., StableBC).

\subsection{Results}

\subsubsection{Intersection (Low-Dimensional)}
Tables~\ref{tab:intersection1_reward} and \ref{tab:intersection2_reward} report \emph{rewards} (higher is better) for the two Intersection scenarios. Overall, \textbf{StableBC} achieves high reward in both scenarios, while \textbf{DeCIL} closes the gap in settings with limited expert data. Because this environment is relatively low-dimensional, \emph{StableBC’s Jacobian-based regularization is tractable.} 
In higher-dimensional tasks, this approach may become more difficult to scale.
% Although \textbf{StableBC} often achieves the best performance overall, \textbf{DeCIL} is competitive when the number of demonstrations is limited (2--10 trajectories). 
% Notably, the Intersection task is relatively low-dimensional, so methods like StableBC, which regularize eigenvalues via Jacobians, can be effective here.

\begin{table*}[!ht]
    \centering
    \resizebox{.75\textwidth}{!}{ 
    \begin{tabular}{lccccc}
    \hline
         & \multicolumn{5}{c}{\textbf{\# of Training Trajectories (Intersection Case 1)}} \\
         & 2 & 5 & 10 & 20 & 50 \\ \hline
        \textbf{BC} 
        & 6.95 $\pm$ 0.52            
        & 11.10 $\pm$ 1.11           
        & 11.89 $\pm$ 1.55           
        & 13.08 $\pm$ 0.66           
        & \textbf{14.05 $\pm$ 0.30}  
        \\ \hline
        \textbf{NoisyBC}     
        & 6.99 $\pm$ 0.64            
        & \textbf{11.58 $\pm$ 1.20}  
        & \textbf{12.21 $\pm$ 1.20}  
        & 13.11 $\pm$ 0.68           
        & 14.00 $\pm$ 0.08           
        \\ \hline
        \textbf{StableBC}
        & \textbf{10.66 $\pm$ 1.09}  
        & \textbf{12.19 $\pm$ 2.75}  
        & \textbf{13.51 $\pm$ 0.42}  
        & \textbf{14.10 $\pm$ 0.55}  
        & \textbf{14.88 $\pm$ 0.33}  
        \\ \hline
        \textbf{DeCIL}
        & \textbf{8.99 $\pm$ 0.37}   
        & 10.76 $\pm$ 2.07           
        & \textbf{12.22 $\pm$ 0.78}  
        & \textbf{13.30 $\pm$ 0.95}  
        & \textbf{14.53 $\pm$ 0.14}  
        \\ \hline
        \textbf{MOReL}
        & \textbf{9.02 $\pm$ 3.64}   
        & \textbf{11.44 $\pm$ 2.68}  
        & 11.64 $\pm$ 2.49           
        & \textbf{14.06 $\pm$ 2.61}  
        & 12.20 $\pm$ 3.64           
        \\
        \hline
    \end{tabular}
    }
    \caption{\textbf{Reward values for Intersection Case 1} (higher is better).
    Although \textbf{StableBC} often ranks the highest overall,
    \textbf{DeCIL} remains competitive in low-data regimes (2--10 trajectories).}
    \label{tab:intersection1_reward}
\end{table*}

\begin{table*}[!ht]
    \centering
    \resizebox{.75\textwidth}{!}{ 
    \begin{tabular}{lccccc}
    \hline
         & \multicolumn{5}{c}{\textbf{\# of Training Trajectories (Intersection Case 2)}} \\
         & 2 & 5 & 10 & 20 & 50 \\ \hline
        \textbf{BC}
        & 7.73 $\pm$ 1.56            
        & 12.08 $\pm$ 0.98           
        & 14.19 $\pm$ 2.40           
        & \textbf{15.49 $\pm$ 0.37}  
        & \textbf{16.60 $\pm$ 0.21}  
        \\ \hline
        \textbf{NoisyBC}
        & 7.47 $\pm$ 2.16            
        & \textbf{13.79 $\pm$ 1.18}  
        & 14.36 $\pm$ 1.70           
        & 15.45 $\pm$ 0.37           
        & 16.19 $\pm$ 0.39           
        \\ \hline
        \textbf{StableBC}
        & \textbf{12.34 $\pm$ 1.43}  
        & \textbf{15.05 $\pm$ 2.95}  
        & \textbf{15.96 $\pm$ 0.58}  
        & \textbf{16.68 $\pm$ 0.70}  
        & \textbf{17.27 $\pm$ 0.43}  
        \\ \hline
        \textbf{DeCIL}
        & \textbf{9.80 $\pm$ 0.27}   
        & \textbf{13.40 $\pm$ 0.67}  
        & \textbf{14.71 $\pm$ 0.98}  
        & \textbf{15.39 $\pm$ 0.81}  
        & \textbf{16.53 $\pm$ 0.34}  
        \\ \hline
        % \textbf{MOReL}
        % & /   & /   & /   & /   & /
        % \\
        % \hline
    \end{tabular}
    }
    \caption{\textbf{Reward values for Intersection Case 2} (higher is better).
    \textbf{StableBC} achieves consistently strong results,
    while \textbf{DeCIL} remains resilient in low-data scenarios. 
    (MOReL results are incomplete for this setup.)}
    \label{tab:intersection2_reward}
\end{table*}

\subsubsection{MetaWorld (Higher-Dimensional)}
We evaluate on four MetaWorld tasks: Button Press, Drawer Close and Coffee Pull. 
Table~\ref{tab:button_press} and Table~\ref{tab:drawer_close} show that on noise-sensitive tasks (Button Press, Drawer Close), the baseline methods degrade quickly with increasing noise, whereas \textbf{DeCIL} remains more robust. 
However, on tasks less sensitive to noise (Coffee Push/Pull), DeCIL provides little benefit over simpler BC-style approaches, which is anticipated.

\begin{table*}[!ht]
    \centering
    \resizebox{\textwidth}{!}{ 
    \begin{tabular}{cccccc}
    \hline
        & 0 & 0.0001 & 0.0005 & 0.001 & 0.002 \\ \hline
        BC & \textbf{3317.14} $\pm$ 208.30 & \textbf{2885.76} $\pm$ 119.74 & 1354.32 $\pm$ 46.86 & 1328.66 $\pm$ 44.65 & 1266.32 $\pm$ 12.27 \\ \hline
        NoisyBC & \textbf{3351.12} $\pm$ 169.07 & 2776.01 $\pm$ 282.97 & 1794.61 $\pm$ 653.78 & 1328.66 $\pm$ 23.29 & 1286.66 $\pm$ 7.78 \\ \hline
        StableBC & 3130.40 $\pm$ 154.91 & 2659.39 $\pm$ 364.43 & 1346.83 $\pm$ 34.36 & 1324.38 $\pm$ 41.17 & 1249.40 $\pm$ 24.47 \\ \hline
        Diffusion & \textbf{3368.96} $\pm$ 135.84 & \textbf{3132.42} $\pm$ 168.63 & 1475.48 $\pm$ 50.93 & 1249.69 $\pm$ 28.09 & 949.99 $\pm$ 32.44 \\ \hline
        DeCIL & 2984.29 $\pm$ 406.87 & \textbf{2971.87} $\pm$ 288.46 & \textbf{2631.77} $\pm$ 422.57 & \textbf{2587.67} $\pm$ 351.73 & \textbf{1861.50} $\pm$ 363.08 \\ \hline
        
    \end{tabular}
    }
    \caption{Rewards for Button Press under different noise levels. 
    \textbf{DeCIL} decays more slowly as noise increases.}
    \label{tab:button_press}
\end{table*}

\begin{table*}[!ht]
    \centering
    \resizebox{\textwidth}{!}{ 
    \begin{tabular}{cccccc}
    \hline
        & 0 & 0.0001 & 0.0005 & 0.001 & 0.002 \\ \hline
        BC & 4252.74 $\pm$ 8.53 & 4198.97 $\pm$ 27.12 & 3922.38 $\pm$ 137.88 & 2123.60 $\pm$ 1164.48 & 823.97 $\pm$ 661.31 \\ \hline
        NoisyBC & 4247.52 $\pm$ 5.64 & 4220.98 $\pm$ 40.88 & 3752.87 $\pm$ 397.34 & 2211.37 $\pm$ 1216.77 & 1069.17 $\pm$ 1300.19 \\ \hline
        StableBC & 4247.52 $\pm$ 8.42 & 4192.03 $\pm$ 52.62 & 3925.91 $\pm$ 131.18 & 2210.85 $\pm$ 1014.67 & 926.93 $\pm$ 822.87 \\ \hline
        Diffusion & 4274.45 $\pm$ 19.38 & 3483.40 $\pm$ 263.43 & 1234.80 $\pm$ 350.43 & 10.57 $\pm$ 2.96 & 241.71 $\pm$ 228.47 \\ \hline
        DeCIL & \textbf{4378.22} $\pm$ 183.30 & \textbf{4405.43} $\pm$ 220.51 & \textbf{4385.55} $\pm$ 244.13 & \textbf{4327.15} $\pm$ 281.49 & \textbf{4001.51} $\pm$ 511.75 \\ \hline
    \end{tabular}
    }
    \caption{Rewards for Drawer Close under different noise levels.
    \textbf{DeCIL} consistently remains above 4000, even with higher noise.}
    \label{tab:drawer_close}
\end{table*}

\begin{table*}[!ht]
    \centering
    \resizebox{\textwidth}{!}{ 
    \begin{tabular}{cccccc}
    \hline
        & 0 & 0.0001 & 0.0005 & 0.001 & 0.002 \\ \hline
        BC & 2764.80 $\pm$ 442.32 & 2811.66 $\pm$ 345.69 & 2791.77 $\pm$ 424.29 & 2642.91 $\pm$ 427.37 & 2501.68 $\pm$ 495.08 \\ \hline
        NoisyBC & 3011.88 $\pm$ 437.37 & 2655.03 $\pm$ 434.13 & 2761.44 $\pm$ 314.87 & 2727.19 $\pm$ 615.97 & 2444.51 $\pm$ 614.48 \\ \hline
        StableBC & 2655.21 $\pm$ 376.88 & 2548.58 $\pm$ 436.41 & 2785.44 $\pm$ 315.25 & 2702.67 $\pm$ 464.76 & 2392.49 $\pm$ 489.06 \\ \hline
        DeCIL & 2150.10 $\pm$ 373.84 & 1818.01 $\pm$ 420.34 & 2624.15 $\pm$ 364.80 & 2213.34 $\pm$ 433.47 & 2365.90 $\pm$ 396.88 \\ \hline
    \end{tabular}
    }
    \caption{Rewards for Coffee Pull under different noise levels. 
    This task is less sensitive to noise; DeCIL does \emph{not} substantially outperform simpler methods.}
    \label{tab:coffee_pull}
\end{table*}

In summary, \textbf{DeCIL} improves robustness in tasks with higher noise sensitivity (Button Press, Drawer Close), but provides little benefit on tasks like Coffee Push/Pull. 
Unlike \textbf{StableBC}, which regularizes based on the Jacobian’s largest eigenvalues, \textbf{DeCIL} avoids the need to compute these terms, making it more scalable in higher-dimensional settings.

% We can see that for the button presses and drawer close tasks, the performance of the baseline method deteriorates rapidly as the noise level increases, as evidenced by the sharp decline in the average rewards. This indicates that these two tasks are highly sensitive to noise, and DeCIL consistently outperforms all the baselines especially in the high noise region.

% For Coffee Pull task, the performance of the baseline methods do not show significant changes as the noise level increases, which indicates that the task is not sensetive to noise. At this point, it is unnecessary to add a denoiser, the results also show that DeCIL does not outperform baseline methods.

% These results indicate that our algorithm improves the model's robustness by denoising, effectively reducing the sensitivity of training tasks to noise. For tasks with lower sensitivity to noise, our algorithm does not outperform the simpler behavior cloning approach, which is anticipated.

\section{Ablation Study}

To verify that DeCIL’s robustness is not merely due to jointly predicting the next state and action, we compare it to a baseline that directly maps the current state to the next state and action without any denoising. We evaluate both methods on two MetaWorld tasks (Button Press and Drawer Close) with increasing noise levels (Figure~\ref{fig:noise_ablation}). 

In both tasks, DeCIL maintains high performance even under significant noise, whereas the baseline’s performance deteriorates rapidly. This confirms that the denoising mechanism is crucial for mitigating noise-induced errors, highlighting its importance for stable control in real-world scenarios.

\section{Conclusion}
\label{sec:conclusion}
We have proposed a novel method to address the covariate shift problem in imitation learning by incorporating a denoising mechanism that enhances the stability of state transitions. Our theoretical analysis demonstrates that the denoising network increases the contraction of the state mapping, ensuring a more stable and reliable policy. Empirical results validate our approach, showing significant improvements over baseline methods. Future work includes extending this framework to high-dimensional state/observation spaces (e.g. images) and investigating its applicability to other sequential decision-making problems.

\section*{Acknowledgements}

The authors would like to thank Heng Yang for discussion at early stage of algorithm prototyping, and Shanghai Qi Zhi Institute for funding and computation resources for this research project.

\nocite{langley00}

\bibliography{example_paper}
\bibliographystyle{icml2024}

%%%%%%%%%%%%%%%%%%%%%%%%%%%%%%%%%%%%%%%%%%%%%%%%%%%%%%%%%%%%%%%%%%%%%%%%%%%%%%%
%%%%%%%%%%%%%%%%%%%%%%%%%%%%%%%%%%%%%%%%%%%%%%%%%%%%%%%%%%%%%%%%%%%%%%%%%%%%%%%
% APPENDIX
%%%%%%%%%%%%%%%%%%%%%%%%%%%%%%%%%%%%%%%%%%%%%%%%%%%%%%%%%%%%%%%%%%%%%%%%%%%%%%%
%%%%%%%%%%%%%%%%%%%%%%%%%%%%%%%%%%%%%%%%%%%%%%%%%%%%%%%%%%%%%%%%%%%%%%%%%%%%%%%
\newpage
\appendix
\onecolumn

\section*{Appendix A: Proof that $L_g < 1$}
\label{appendix:proof}

\subsection*{Objective}

We aim to prove that the denoising policy network \( g \), trained with the denoising objective, has a Lipschitz constant \( L_g < 1 \) with respect to its input \( y \) (the noisy next state). This ensures that \( g \) acts as a contraction mapping, enhancing the stability of the state transition from \( x_t \) to \( x_{t+1} \).

\subsection*{Definitions and Assumptions}

Let:

\begin{itemize}
    \item \( g: \mathcal{X} \times \mathcal{X} \rightarrow \mathcal{X} \times \mathcal{A} \) be the denoising network mapping the current state \( x_t \) and a noisy next state \( y \) to a denoised next state \( \hat{x}_{t+1} \) and action \( \hat{a}_t \):
    \[
        [\hat{x}_{t+1}, \hat{a}_t] = g(x_t, y).
    \]
    \item The noisy input \( y \) is defined as:
    \[
        y = x_{t+1} + \eta,
    \]
    where \( \eta \) is additive noise sampled from a zero-mean Gaussian distribution \( \eta \sim \mathcal{N}(0, \sigma^2 I) \).
    \item The denoising objective is:
    \[
        \mathcal{L}_{\text{denoise}} = \mathbb{E}_{x_{t+1}, \eta} \left[ \left\| g(x_t, x_{t+1} + \eta) - x_{t+1} \right\|^2 \right].
    \]
    \item We assume \( g \) is differentiable with respect to \( y \) and has a Lipschitz constant \( L_g \) in \( y \).
\end{itemize}

Our goal is to show that \( L_g < 1 \).

\subsection*{Proof}

\paragraph{Step 1: Taylor Expansion of \( g \)}

For small noise \( \eta \), we perform a first-order Taylor expansion of \( g \) around \( y = x_{t+1} \):
\[
    g(x_t, x_{t+1} + \eta) \approx g(x_t, x_{t+1}) + J_g \eta,
\]
where:
\begin{itemize}
    \item \( g(x_t, x_{t+1}) = x_{t+1} \) (since \( g \) is trained to output the clean next state when there is no noise).
    \item \( J_g = \dfrac{\partial g}{\partial y} \bigg|_{y = x_{t+1}} \) is the Jacobian matrix of \( g \) with respect to \( y \) evaluated at \( y = x_{t+1} \).
\end{itemize}

\paragraph{Step 2: Approximate the Denoising Loss}

Substituting the Taylor expansion into the denoising loss:
\[
    \mathcal{L}_{\text{denoise}} \approx \mathbb{E}_{\eta} \left[ \left\| g(x_t, x_{t+1} + \eta) - x_{t+1} \right\|^2 \right] = \mathbb{E}_{\eta} \left[ \left\| J_g \eta \right\|^2 \right].
\]

\paragraph{Step 3: Compute the Expectation}

Since \( \eta \sim \mathcal{N}(0, \sigma^2 I) \), we have:
\[
    \mathbb{E}_{\eta} \left[ \eta \eta^\top \right] = \sigma^2 I.
\]

Therefore,
\begin{align*}
    \mathcal{L}_{\text{denoise}} &\approx \mathbb{E}_{\eta} \left[ \eta^\top J_g^\top J_g \eta \right] \\
    &= \operatorname{Tr}\left( J_g^\top J_g \mathbb{E}_{\eta} \left[ \eta \eta^\top \right] \right) \\
    &= \sigma^2 \operatorname{Tr}\left( J_g^\top J_g \right) \\
    &= \sigma^2 \left\| J_g \right\|_F^2,
\end{align*}
where \( \left\| J_g \right\|_F \) is the Frobenius norm of the Jacobian \( J_g \).

\paragraph{Step 4: Minimization of the Denoising Loss}

Minimizing \( \mathcal{L}_{\text{denoise}} \) with respect to \( g \) is equivalent to minimizing \( \left\| J_g \right\|_F^2 \):
\[
    \min_{g} \mathcal{L}_{\text{denoise}} \quad \Leftrightarrow \quad \min_{g} \left\| J_g \right\|_F^2.
\]

This implies that the training process encourages the Jacobian \( J_g \) to have a small Frobenius norm.

\subsection*{Discussion}

The key intuition behind this proof is that the denoising objective inherently penalizes the sensitivity of \( g \) to changes in its input \( y \). By minimizing the reconstruction error caused by noise \( \eta \), the network is encouraged to map nearby inputs (i.e., \( y_1 \) and \( y_2 \) that are close in norm) to outputs that are even closer, due to the contraction property (since \( L_g < 1 \)).

This contraction property is crucial for enhancing the stability of the state transition mapping. It ensures that errors introduced in the predicted next state \( \hat{x}_{t+1} \) are progressively reduced by \( g \), mitigating the compounding of errors over time and effectively addressing the covariate shift problem.


% You can have as much text here as you want. The main body must be at most $8$ pages long.
% For the final version, one more page can be added.
% If you want, you can use an appendix like this one.  

% The $\mathtt{\backslash onecolumn}$ command above can be kept in place if you prefer a one-column appendix, or can be removed if you prefer a two-column appendix.  Apart from this possible change, the style (font size, spacing, margins, page numbering, etc.) should be kept the same as the main body.
%%%%%%%%%%%%%%%%%%%%%%%%%%%%%%%%%%%%%%%%%%%%%%%%%%%%%%%%%%%%%%%%%%%%%%%%%%%%%%%
%%%%%%%%%%%%%%%%%%%%%%%%%%%%%%%%%%%%%%%%%%%%%%%%%%%%%%%%%%%%%%%%%%%%%%%%%%%%%%%

\end{document}